\documentclass[runningheads]{llncs}
\usepackage{mathtools}
\usepackage{bm}
\usepackage{nicefrac}
\usepackage{microtype}

\usepackage{color,xcolor}
\usepackage{epsfig}
\usepackage{graphicx}

\usepackage{adjustbox}
\usepackage{array}
\usepackage{booktabs}
\usepackage{colortbl}
\usepackage{hyperref}
\usepackage{wrapfig}
\usepackage{hhline}
\usepackage{multirow}
\usepackage{subcaption}
\usepackage[size=small]{caption}

\usepackage{changepage}
\usepackage{extramarks}
\usepackage{fancyhdr}
\usepackage{lastpage}
\usepackage{setspace}
\usepackage{soul}
\usepackage{xspace}

\usepackage{url}

\usepackage{algpseudocode}
\usepackage{algorithmicx}
\usepackage[ruled]{algorithm2e}
\usepackage{enumerate}
\usepackage{enumitem}  %
\usepackage{makecell}

\newcolumntype{L}[1]{>{\raggedright\let\newline\\\arraybackslash\hspace{0pt}}m{#1}}
\newcolumntype{C}[1]{>{\centering\let\newline\\\arraybackslash\hspace{0pt}}m{#1}}
\newcolumntype{R}[1]{>{\raggedleft\let\newline\\\arraybackslash\hspace{0pt}}m{#1}}

\newcommand{\sect}[1]{Section~\ref{#1}}
\newcommand{\sectapp}[1]{Appendix~\ref{#1}}

\newcommand{\eqn}[1]{Equation~\eqref{#1}}
\newcommand{\fig}[1]{Fig.~\ref{#1}}
\newcommand{\tbl}[1]{Table~\ref{#1}}

\newcommand{\ignore}[1]{}

\DeclareMathAlphabet{\mathbfit}{OML}{cmm}{b}{it}

\makeatletter
\DeclareRobustCommand\onedot{\futurelet\@let@token\@onedot}
\def\@onedot{\ifx\@let@token.\else.\null\fi\xspace}

\def\etal{et al\onedot}

\makeatother

\definecolor{MyDarkBlue}{rgb}{0,0.08,1}
\definecolor{MyAqua}{rgb}{0,0.7,0.7}
\definecolor{MyDarkGreen}{rgb}{0.02,0.6,0.02}
\definecolor{MyDarkRed}{rgb}{0.8,0.02,0.02}
\definecolor{MyDarkOrange}{rgb}{0.40,0.2,0.02}
\definecolor{MyPurple}{RGB}{111,0,255}
\definecolor{MyRed}{rgb}{1.0,0.0,0.0}
\definecolor{MyGold}{rgb}{0.75,0.6,0.12}
\definecolor{MyDarkgray}{rgb}{0.66, 0.66, 0.66}

\SetKwFor{For}{for }{}{}

\newcommand{\modelfull}{Video Extrapolation in Space and Time\xspace}
\newcommand{\model}{VEST\xspace}
\newcommand{\nvs}{NVS\xspace}
\newcommand{\vp}{VP\xspace}
\newcommand{\na}{N/A\xspace}

\newcommand{\myparagraph}[1]{\vspace{-5pt}\paragraph{#1}}

\usepackage{graphicx}

\usepackage{tikz}
\usepackage{comment}
\usepackage{amsmath,amssymb} %
\usepackage{color}

\usepackage[accsupp]{axessibility}  %

\begin{document}
\pagestyle{headings}
\mainmatter
\def\ECCVSubNumber{4275}  %

\title{Video Extrapolation in Space and Time} %

\titlerunning{Video Extrapolation in Space and Time}
\author{Yunzhi Zhang \and Jiajun Wu
}
\authorrunning{Y. Zhang and J. Wu}
\institute{Stanford University\\
\email{\{yzzhang,jiajunwu\}@cs.stanford.edu}
}

\maketitle

\begin{abstract}
Novel view synthesis (\nvs) and video prediction (\vp) are typically considered disjoint tasks in computer vision. However, they can both be seen as ways to observe the spatial-temporal world: \nvs aims to synthesize a scene from a new point of view, while \vp aims to see a scene from a new point of time. These two tasks provide complementary signals to obtain a scene representation, as viewpoint changes from spatial observations inform depth, and temporal observations inform the motion of cameras and individual objects. 
Inspired by these observations, we propose to study the problem of \modelfull (\model). We propose a model that leverages the self-supervision and the complementary cues from both tasks, while existing methods can only solve one of them.
Experiments show that our method achieves performance better than or comparable to several state-of-the-art \nvs and \vp methods on indoor and outdoor real-world datasets. \footnote{
Project page: \url{https://cs.stanford.edu/~yzzhang/projects/vest/}.
}

\end{abstract}

\begin{figure*}[t]
    \centering
    \includegraphics[width=\linewidth]{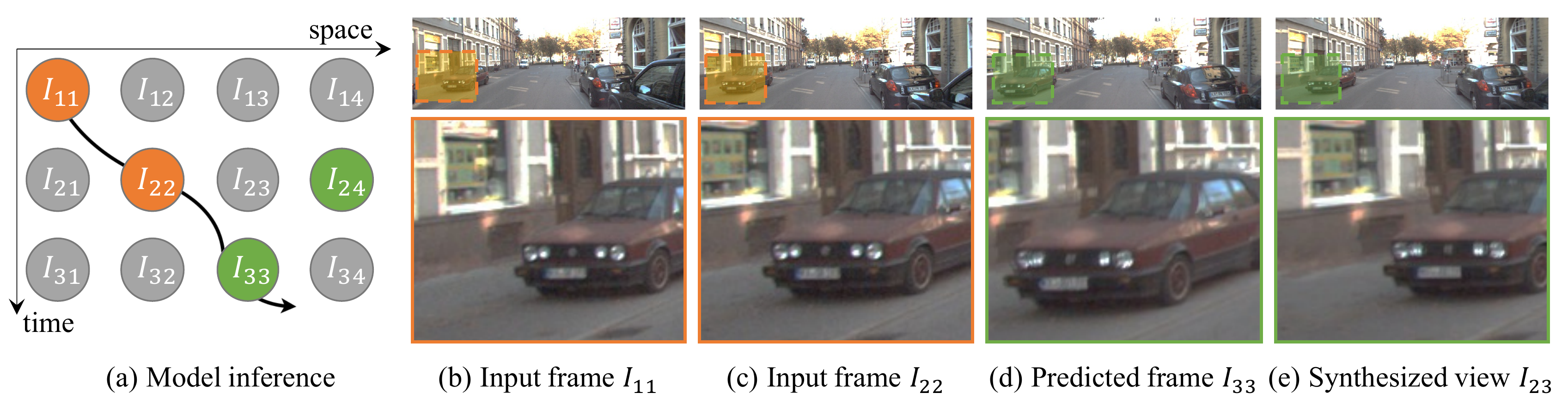}
    \vspace{-10pt}
    \caption{We propose the task of \modelfull(\model) that exploits both spatial and temporal consistency in video data. 
    (a) During inference time, the model takes in two consecutive frames as inputs, colored in orange, and outputs images extrapolated along the temporal axis for future frame prediction and along the spatial axis for novel view synthesis, colored in green. The black curve denotes the camera trajectory.
    (b) and (c) are examples of model inputs; (d) and (e) are model outputs.
    Images from the bottom row are zoom-in views.
    }
    \label{fig:teaser}
    \vspace{-10pt}
\end{figure*}

\section{Introduction}
\label{sec:intro}

Novel view synthesis (\nvs) and video prediction (\vp) are both widely studied computer vision tasks. The former extrapolates a scene to a different camera viewpoint, while the latter extrapolates to a future timestamp. 
\nvs focuses on the scene geometry revealed from discrete camera positions, whereas \vp extracts information from the moving trajectories of both cameras and objects. The self-supervision signals from these two tasks can be jointly used to extract a scene representation. To this end, we call attention to the problem of \modelfull (\model) that considers both tasks.

We solve the problem of \model by first developing a representation that incorporates both the spatial and temporal consistency from video data as inductive biases. We generalize Multiplane Images (MPIs)~\cite{zhou2018stereo}, which is originally a layered representation that decomposes images into RGBA planes, by additionally parameterizing the flow field of each plane in order to model the temporal dynamics. Images from a future timestamp and a novel viewpoint can be rendered from flow-based and homography-based warping, respectively.

Compared with previous MPI-based \nvs approaches~\cite{flynn2019deepview,srinivasan2019pushing,zhou2018stereo,tucker2020single}, our generalized MPI representation leverages learning signals derived from both the spatial and temporal coherence in video data, while previous \nvs methods utilize frame tuples randomly sampled from the training video sequences without considering the temporal information. 
Compared with other optical-flow-based \vp methods~\cite{gao2019disentangling}, our method predicts the motion field individually for each MPI plane instead of the full scene. Since each MPI plane captures a relatively simple structure, we can effectively estimate the motion field of each plane with affine transformations~\cite{wang1993layered}. 

We instantiate a model that performs the spatial-temporal extrapolation with the generalized MPI representation. As shown in \fig{fig:teaser}, our model predicts MPI planes from monocular inputs and leverages historical frames for motion inference. 
Experiments show that our problem formulation and model are generally applicable to diverse scenarios: indoor and outdoor scenes, and videos taken by a single and multiple static or moving camera(s). 
Our method achieves favorable performance on {\it both} tasks compared to baselines designed for either.

Our main contributions are as follows:
\begin{itemize}
    \item We propose \model as a self-supervision task for learning a generalized MPI representation from video data.
    \item We instantiate a model that learns the proposed representation for simultaneous video extrapolation in space and time.
    \item Our \model model produces realistic results in both space and time extrapolation on a diverse range of datasets.
\end{itemize}

\begin{figure*}[t]
    \centering
    \includegraphics[width=\linewidth]{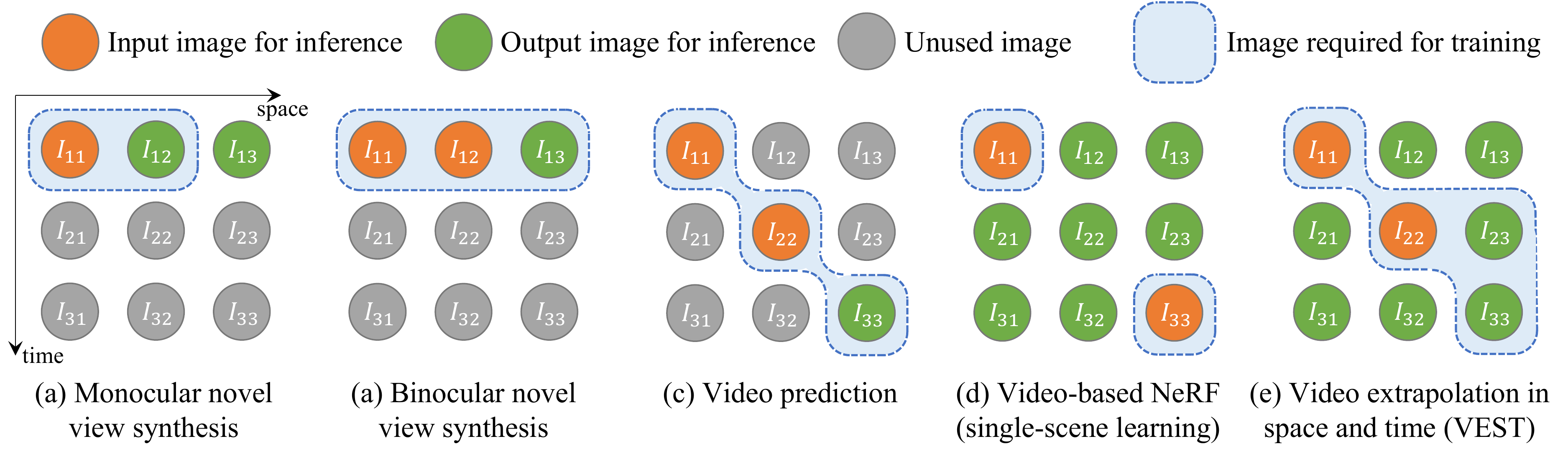}
    \vspace{-10pt}
    \caption{Overview of vision tasks related to our work. $I_{ij}$ denotes the RGB frame for the $i$-th timestamp in a video sequence taken from the $j$-th camera viewpoint. We restrict $i, j \in \{1, 2, 3\}$ for illustration. In (a) and (b), monocular or binocular \nvs methods only extrapolate in space, while in (c), \vp methods only extrapolate in time. Video-based NeRF methods in (d) train on a collection of images for each scene, and perform interpolation instead of extrapolation for inference.
    Finally, (e) illustrates the problem we focus on in this work. Given historical frames, the task consists of both predicting the future and extrapolating to a novel viewpoint. During inference, novel views $I_{21}, I_{23}$ can be synthesized the same way as training time, while $I_{22}, I_{23}$ can be inferred by duplicating $I_{11}$ as input (\model-single from \ref{section:multiple}).
    }
    \vspace{-10pt}
    \label{fig:overview}
\end{figure*}

\section{Related Works}
\label{section:related}

\paragraph{Novel view synthesis.} Synthesizing novel views based on 2D images is a challenging problem, as it requires the reasoning of the 3D structure of the perceptual world. 
As shown in~\fig{fig:overview}(a) and~\fig{fig:overview}(b), monocular view synthesis approaches~\cite{tucker2020single,wiles2020synsin,lai2021video} take $I_{11}$ as inputs and outputs the synthesized image for a query view $j \neq 1$. Stereo view synthesis approaches~\cite{zhou2018stereo} are similar but take in two input views $I_{11}, I_{12}$. 
Most of these approaches synthesize novel views based on camera parameters estimated from Structure-from-Motion (SfM) techniques, except for Lai~\etal~\cite{lai2021video} which uses camera parameters predicted by the model. These methods focus exclusively on spatial extrapolation and do not take the temporal axis into account. 
Previous works~\cite{zhou2016view,zhou2018stereo,shih20203d,tulsiani2018layer} use layered representation and apply per-layer warping to obtain the novel view based on camera poses. 
SynSin~\cite{wiles2020synsin} and Worldsheet~\cite{hu2021worldsheet} predict the depth from a monocular image input and warps a 3D representation of the scene in the features space and pixel space, respectively, to synthesize novel views. 
In comparison, our proposed VEST encapsulates NVS but also has the ability to perform temporal extrapolation.

Closer to ours, Lin~\etal~\cite{lin2021deep} and Yoon~\etal~\cite{yoon2020novel} tackle the problem of NVS for dynamic scenes. Lin~\etal~\cite{lin2021deep} addresses the temporal inconsistency of the MPI representation when applied to scenes with moving objects by identifying the error-prone regions with a learned 3D mask volume. 
There are three key differences between our model and Lin~\etal.
First, they require two images from synchronized stereo cameras as inputs during inference, while our method takes in two consecutive frames from a single camera.
Second, they assume a static camera and static background while we do not. 
Third, they rely on the static background computed from the full video sequences from two source views, while ours directly predicts the future video sequence based on two input frames. 
Yoon~\etal~\cite{yoon2020novel} proposes to fuse the depth estimated from a single view and multiple views with a depth fusion network, and predicts the novel view on top. It requires segmentation labels and optical flow inputs, while our method is fully self-supervised.

\myparagraph{Video prediction.} 
Video prediction methods typically take in two or more frames taken by one camera and only extrapolate along the temporal axis as shown in~\fig{fig:overview}(c). 
Deep neural networks have been widely used in video prediction~\cite{ranzato2014video,srivastava2015unsupervised,shi2015convolutional,gao2019disentangling}. Previous works studying temporal extrapolation for videos include hallucination-based methods~\cite{lotter2016deep,villegas2017decomposing} and warping-based methods~\cite{liu2017video}. More recently, Wang~\etal~\cite{wang2022predrnn} propose PredRNN-V2, integrating convolutional recurrent units with a pair of decoupled memory
cells. 
Several methods also proposed to decompose high-dimensional videos into object-centric~\cite{wu2020future} or semantic-aware~\cite{bei2021learning} regions, and model the motion of each region independently. These methods consider only monocular input and often require models pre-trained on large datasets for semantic and depth information, while our method factorizes videos into depth-aware components by using the geometric cue from view synthesis, and it requires no external supervision. 

\myparagraph{Dynamic neural radiance fields.} Neural Radiance Fields (NeRF)~\cite{mildenhall2020nerf} shows impressive results in synthesizing novel views with high fidelity. Not surprisingly, many follow-up papers have attempted to extend it to video for image synthesis in novel space and time point~\cite{du2021neural}. Our method differs from these methods in two ways. First, though on static scenes, methods like PixelNeRF can already learn a neural scene representation conditioned on one or few input images~\cite{yu2021pixelnerf,park2021nerfies,xian2021space,li2021neural,pumarola2021d,tretschk2021non}, all video-based NeRF methods still employ single-video learning and need to be re-trained for every new scene, while our method performs 4D synthesis conditioned on the input images. Second, NeRF-based methods require hundreds of images for training, and essentially perform interpolation among input time- and viewpoints as shown in~\fig{fig:overview}(d), while our method focuses on video extrapolation.

\myparagraph{Layered representations.}
The idea of decomposing images into RGBA layers is effective in multiple problem domains. For view synthesis, Shade~\etal~\cite{shade1998layered} proposed layered depth images to represent a scene with multi-layer depth and color images. Viewers can then see the scene from different points in space. For video prediction, there has also been a line of work inspired by the classic research on layered motion representations~\cite{wang1993layered}. A notable example is a recent work by Lu~\etal~\cite{lu2020,lu2021omnimatte} on the problem of video manipulation, where they decompose videos into semantic-aware RGBA layers. Our layered representation builds upon all these ideas and aims to tackle both \nvs and \vp.

\section{Method}
\subsection{Multiplane Images}
\label{MPI}
Before introducing the representation we use, we first review the classical Multiplane Images (MPIs)~\cite{zhou2018stereo}. 
An image $I \in \mathbb{R}^{H \times W \times 3}$ is represented by $D$ RGBA planes,$\{(c_i, \alpha_i)\}_{i=1}^D$, where $c_i \in \mathbb{R}^{H\times W \times 3}$ are RGB values and $\alpha_i \in \mathbb{R}^{H\times W \times 1}$ are alpha values. Each plane corresponds to fixed depth $d_i$. The plane is fronto-parallel to the camera and can be written as $\boldsymbol{n}^T \boldsymbol{x} - d_i = 0$, where $\boldsymbol{n} = \begin{bmatrix}0,0,1\end{bmatrix}^T$ is the normal vector of the plane. 

Let $(R, \boldsymbol{t})$ be the rotation and translation matrix from target to source view, and $K, K^\prime$ be the camera intrinsics for source and target views. The transformation for the $i$-th plane from target to source view, denoted as $\mathcal{W}_i^{R, \boldsymbol{t}, K, K^\prime}$, is defined as
\begin{equation}
\label{mpi_warp}
    \begin{bmatrix}u \\ v \\ 1\end{bmatrix} 
    \sim \mathcal{W}_i^{R, \boldsymbol{t}, K, K^\prime} \begin{bmatrix}u^\prime \\v^\prime \\1\end{bmatrix}
    := K \bigl( R - \frac{\boldsymbol{t} \boldsymbol{n}^T}{d_i} \bigr) (K^\prime)^{-1} \begin{bmatrix}u^\prime \\v^\prime \\1\end{bmatrix}, 
\end{equation}
where $(u, v)$ and $(u^\prime, v^\prime)$ are the coordinates from the source and target views, respectively. 

The MPI representation for the target view $\{\left(c^\prime_i, \alpha^\prime_i\right)\}_{i=1}^D$ is computed as 
\begin{align}
    \label{warp_color}
    c^\prime_{u^\prime, v^\prime} &= c_{u, v},\\
    \label{warp_alpha}
    \alpha^\prime_{u^\prime, v^\prime} &= \alpha_{u, v}.
\end{align}
Here $(u, v)$ are sampled according to \eqn{mpi_warp}. Similar to Zhou~\etal~\cite{zhou2018stereo}, we apply bilinear sampling from the neighboring grid corners when $(u, v)$ is not aligned with the coordinate grid. 

Finally, an MPI renderer synthesizes the target-view image with 
\begin{equation}
\label{mpi_composition}
    \hat{I}^\prime = \sum_{i=1}^D c_i^\prime \alpha_i^\prime \prod_{j=i+1}^{D} (1 - \alpha_j^\prime).
\end{equation}

\subsection{Generalized Multiplane Images}
\label{DMPI}

We generalize the MPI representation to model the motion from image $I = I_t$ to $I^\prime = I_{t+1}$, the next frame in a video sequence. 
Note that different from Section~\ref{MPI}, here we use $I^\prime$ to refer to the frame for timestamp $t+1$ which is not necessarily from a different camera viewpoint.

Formally, the generalized MPI representation for image $I$ is denoted as $\{(c_i, \alpha_i, \mathcal{T}_i)\}_{i=1}^D$, where $\mathcal{T} = \mathcal{T}_i$ is an operator defined by
\begin{equation}\label{dmpi_warp}
\begin{bmatrix}u \\ v\end{bmatrix} \sim \mathcal{T}\begin{bmatrix}u^\prime \\v^\prime \end{bmatrix} := 
\begin{bmatrix}u^\prime + \Delta u^\prime \\v^\prime + \Delta v^\prime \end{bmatrix}.
\end{equation}

Let $\theta$ be a parameterization for $\mathcal{T}$. When $\mathcal{T}$ is fully parameterized, $\theta \in \mathbb{R}^{H \times W \times 2}$ is the pixel displacement field, and $\Delta u^\prime = \theta_{u^\prime, v^\prime, 0}$, $\Delta v^\prime = \theta_{u^\prime, v^\prime, 1}$ correspond to the number of pixels to be shifted in the $u$, $v$ coordinates, respectively. However, since each MPI plane has a relatively simple structure, in practice we restrict $\mathcal{T}$ to be in the class of affine transformations. $\mathcal{T}$ can now be parameterized by $\theta\in \mathbb{R}^{2 \times 3}$, and
\begin{equation}
    \mathcal{T}\begin{bmatrix}u^\prime \\v^\prime \end{bmatrix} = \begin{bmatrix}\theta_{1, 1}& \theta_{1, 2}& \theta_{1, 3} \\ \theta_{2, 1}&\theta_{2, 2}&\theta_{2, 3}\end{bmatrix}\begin{bmatrix}u^\prime \\v^\prime \\1\end{bmatrix}.
\end{equation}
We denote the generalized MPI representation for image $I$ as
\begin{equation}
\label{generalized_mpi}
    \{(c_i, \alpha_i, \theta_i)\}_{i=1}^D. 
\end{equation}
Finally, we predict the next frame $\hat{I}^\prime$ based on~\eqn{warp_color}-\eqref{mpi_composition}.

\begin{figure*}[t]
    \centering
    \includegraphics[width=\linewidth]{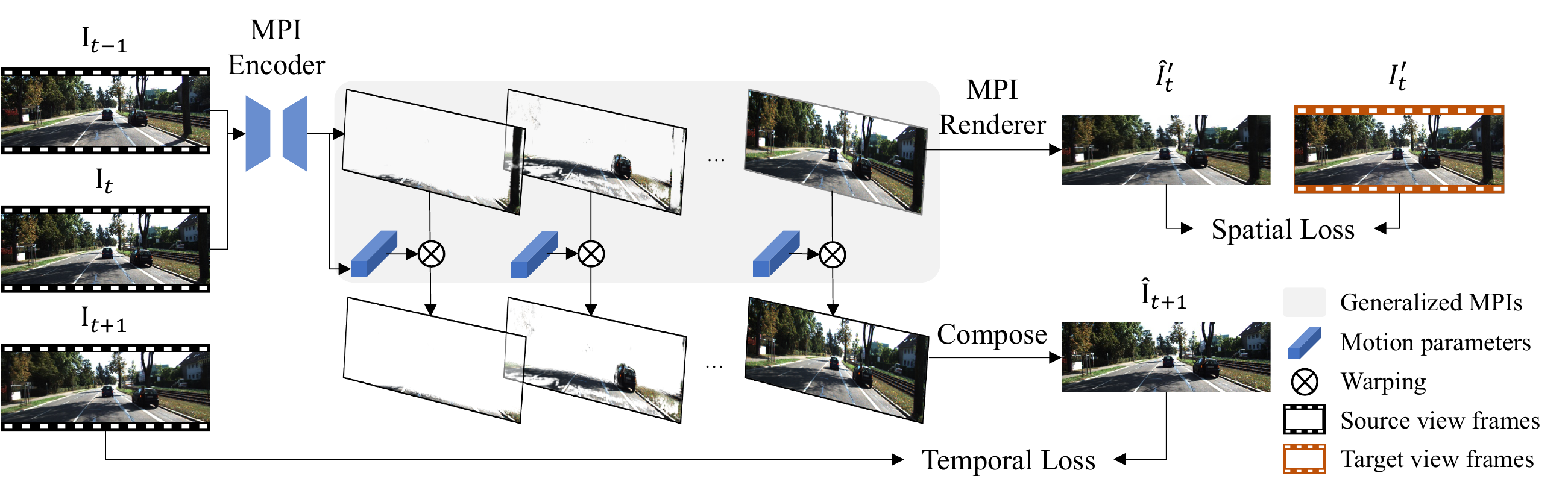}
    \caption{Model architecture. The MPI encoder receives monocular frames $I_{t-1}$ and $I_{t}$ as inputs and outputs the generalized MPI representation for $I_{t}$, which consists of $D$ RGBA planes and the motion parameters for each plane. Then the MPI renderer renders the target view image $I_t^\prime$ based on camera parameters and the RGBA planes. The next-frame prediction $I_{t+1}$ is generated by first warping each RGBA plane with motion parameters, and then composing the planes. The training objective is to match $\hat{I}_t^\prime$ and $\hat{I}_{t+1}$ with the ground truth. Images are licensed under CC BY-NC-SA 3.0.}
    \vspace{-10pt}
    \label{fig:model}
    
\end{figure*}

\subsection{\modelfull}

We now introduce our \model model that predicts the generalized MPIs given monocular video frames, and performs spatial and temporal extrapolation. An overview of our model is shown in~\fig{fig:model}. 

\myparagraph{Training.}
The model takes in two consecutive frames from a video sequence, $I_{t-1}$ and $I_{t}$, and outputs the generalized MPI representation (\eqn{generalized_mpi}) for $I_t$. 
The target-view image $I_t^\prime$ is synthesized following \sect{MPI}, and the next-frame prediction following \sect{DMPI}. 

\myparagraph{Inference.}
During inference, with inputs $I_{t-1}$ and $I_{t}$, the model can be queried to extrapolate to other space-time coordinates. Future frames with longer horizon $I_{t+2}, I_{t+3}, \cdots$ can be inferred by iteratively forwarding the model in an autoregressive manner. 

Even when there is only one input frame $I_t$ available, our model can synthesize the novel view $I_t^\prime$ by having $I_t$ replicated twice as inputs and still produce realistic \nvs results, corresponding to VEST-single from \tbl{table:kitti} and \tbl{table:estate} as we will show later.

\myparagraph{Losses.}
The training loss is the sum of spatial and temporal extrapolation errors, namely 
\begin{equation}
    \mathcal{L}^\text{total} = 
\mathcal{L}^\text{space}(\hat{I}_t^\prime, I_t^\prime) 
+ \mathcal{L}^\text{time}(\hat{I}_{t+1}, I_{t+1}),
\end{equation}
where
\begin{align}
\label{eqn:spatial}
    \mathcal{L}^\text{space} &= 
\lambda_{1}^\text{space}\mathcal{L}_1 + \lambda_\text{perc}^\text{space} \mathcal{L}_\text{perc},\\
\mathcal{L}^\text{time} &= 
\lambda_{1}^\text{time}\mathcal{L}_1 + \lambda_\text{perc}^\text{time} \mathcal{L}_\text{perc}.
\end{align}
$\mathcal{L}_1$ is the $\ell_1$ loss, and $\mathcal{L}_\text{perc}$ is the perceptual loss using pretrained VGG-19~\cite{simonyan2014very} features. 

\subsection{Implementation Details}

We adopt a model architecture similar to Tucker and Snavely~\cite{tucker2020single}, specified in \sectapp{appendix:architecture}. The network outputs a tensor in $\mathbb{R}^{D\times H\times W\times 7}$, which is split channelwise into $f^\alpha \in \mathbb{R}^{D \times H \times W \times 1}$ for alpha values and $f^\theta\in \mathbb{R}^{D \times H\times W \times 6}$ for motion parameters. We set the RGB values for each plane ($c_i$ from \eqn{generalized_mpi}) to be the RGB values of the source image.
The final motion parameter $\theta$ for each MPI plane is computed as a weighted spatial average of $f^\theta$:
\begin{align}
\label{eqn:mpi_weights_src}
    &w_i = \alpha_i \prod_{j=i+1}^{D} (1 - \alpha_j), \\
    &\theta_i = \text{SpatialAverage}(w_i \otimes f^\theta),
\end{align}
where $\otimes$ denotes the element-wise multiplication. Note that $w_i$ are the same weights for RGB values used in \eqn{mpi_composition}.

Camera parameters are estimated with SfM~\cite{schoenberger2016sfm,schoenberger2016mvs} and are used to render novel views. Since SfM models have ambiguous depth scales, we compute a depth scale factor $\sigma$ similar to Tucker and Snavely~\cite{tucker2020single} such that MPI planes are associated with scaled depth values $\{\sigma d_i\}_{i=1}^D$ instead of $\{d_i\}_{i=1}^D$. Here $\sigma$ is computed to minimize the log-squared error of the predicted depth map $\hat{Z}$ and a set of sparse 3D points $P_s$,
\begin{equation}\label{depth_scale}
    \sigma = \exp{\left[\frac{1}{\mid P_s\mid}\sum_{(u, v, z)\in P_s}(\ln z - \ln\hat{Z}_{u, v})\right]},
\end{equation}
where the depth map is computed with \eqn{eqn:mpi_weights_src} and 
\begin{equation}\label{depth_composition}
    \hat{Z} = \sum_{i=1}^D d_i w_i. %
\end{equation}
For each element $(u, v, z) \in P_s$, $(u, v)$ is the pixel coordinate in the source image, and $z$ is the depth value of the corresponding 3D point. We run COLMAP~\cite{schoenberger2016sfm} to obtain $P_s$ for each video sequence.

\section{Experiments}

\begin{figure}[t]
    
\begin{minipage}[l]{0.45\textwidth}
    \centering
    \vspace{-10pt}
    \includegraphics[width=\linewidth]{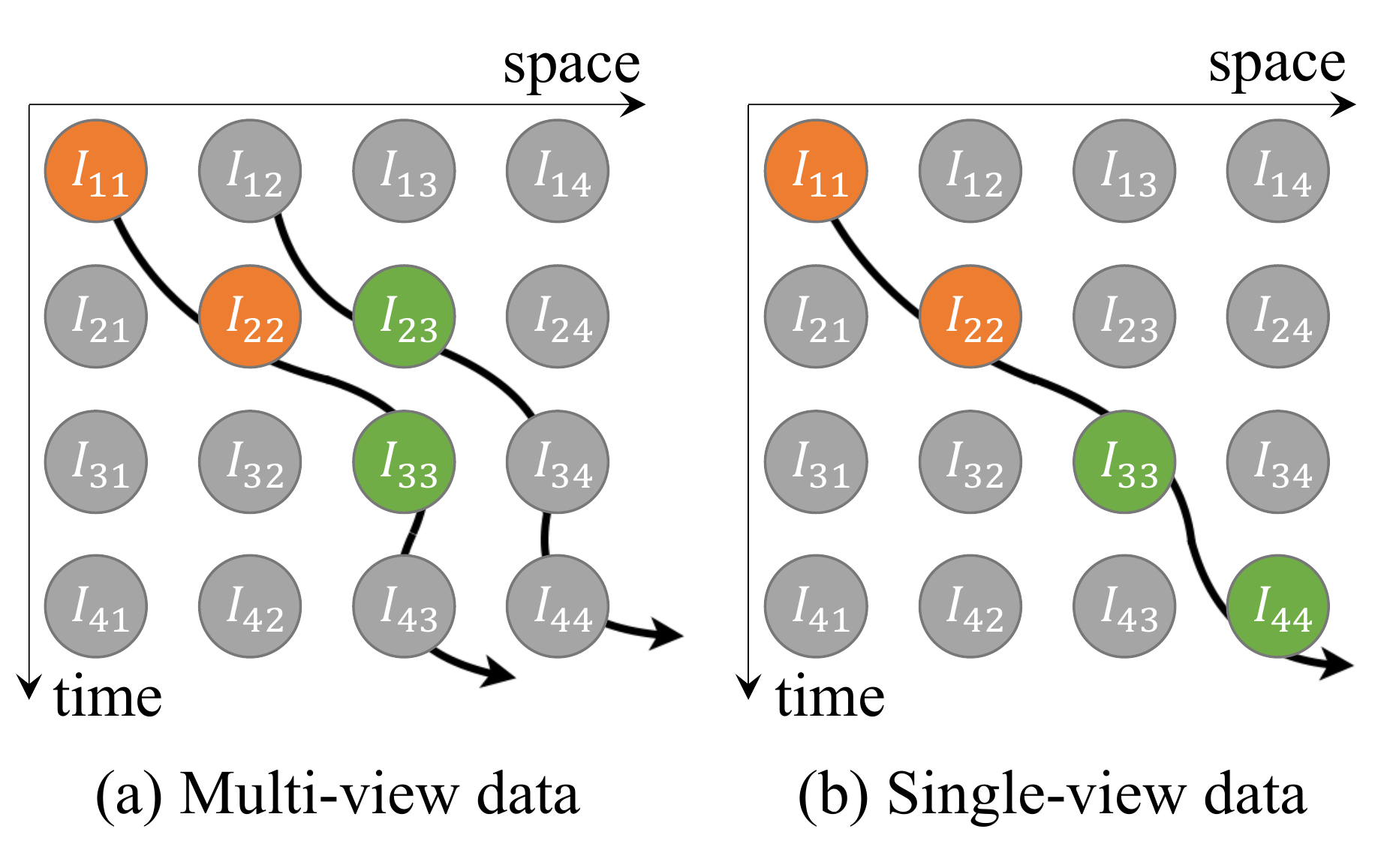}
\end{minipage}\hfill
\begin{minipage}[r]{0.5\textwidth}
    \caption{Training setups. Training inputs are colored in orange and outputs are in green. Black curves denote camera trajectories. Our method can be trained on multi-view or single-view datasets. $I_{33}$ is the ground truth for future frame prediction. The NVS supervision comes from a stereo frame when available ($I_{23}$ in (a)), or from a temporally nearby frame otherwise ($I_{44}$ in (b)).
    }
    \label{fig:setups}
\end{minipage}
\vspace{-15pt}

\end{figure}
\begin{table}[t]
\centering\scriptsize
\setlength{\tabcolsep}{2pt}
\begin{tabular}{clccccccccc}
\toprule
\multirow{3}{*}{Split} & \multirow{3}{*}{Method} & \multicolumn{3}{c}{Extrapolation in Space} & \multicolumn{3}{c}{Extrapolation in Time} \\ \cmidrule(l){3-8} 
  &  & \multirow{2}{*}{LPIPS$\downarrow$} & \multirow{2}{*}{SSIM$\uparrow$} & \multirow{2}{*}{PSNR$\uparrow$} & \multicolumn{3}{c}{LPIPS-AlexNet$\downarrow$} 
 \\ \cmidrule(l){6-8}
 &  &  &  &  & $t+1$ & $t+3$ & $t+5$ \\
 \midrule
\multirow{7}{*}{
\makecell{LDI~\cite{tulsiani2018layer} split\\ train 256 $\times$ 768 \\ test 128 $\times$ 384}
} 
& MINE~\cite{limine}-32* & 0.112 & 0.822 & 21.4 & \multicolumn{3}{c}{\na} \\
 & MINE~\cite{limine}-64* & 0.108 & 0.820 & 21.3 & \multicolumn{3}{c}{\na} \\ \cmidrule(l){2-8} 
 & LDI~\cite{tulsiani2018layer} & \na& 0.572 & 16.5 & \multicolumn{3}{c}{\na} \\
 & Tucker~\etal~\cite{tucker2020single} & \na & 0.733 & 19.5 & \multicolumn{3}{c}{\na} \\
 & PredRNN-V2~\cite{wang2022predrnn} & \multicolumn{3}{c}{\na} & 0.3085 & 0.4573 & 0.5422 
 \\
 
 & \model-no-perc & 0.111 & 0.819 & 21.4 & 0.1129 & 0.2902 & 0.3943
\\

 & \model-single & 0.091 & 0.816 & 21.2 & \multicolumn{3}{c}{\na} \\
 & \model (ours) & \textbf{0.085} & \textbf{0.825} & \textbf{21.6} & \textbf{0.1154} & \textbf{0.2881} & \textbf{0.3911} 
 \\ \midrule
 
\multirow{5}{*}{
\makecell{
LDI~\cite{tulsiani2018layer} split\\
train 128 $\times$ 384\\
test 128 $\times$ 384
}
}
& MINE~\cite{limine}* & 0.112 & 0.828 & 21.9 & \multicolumn{3}{c}{\na} \\\cmidrule(l){2-8} 
& MINE~\cite{limine} & 0.129 & 0.812 & \textbf{21.4} & \multicolumn{3}{c}{\na} \\
 & PredRNN-V2~\cite{wang2022predrnn} & \multicolumn{3}{c}{\na} & 0.2153 & 0.3946 & 0.4984 
 \\
 & \model-single & 0.105 & 0.806 & 20.7 & \multicolumn{3}{c}{\na} \\
 & \model (ours) & \textbf{0.097} & \textbf{0.818} & 21.1 & \textbf{0.0798} & \textbf{0.2348} & \textbf{0.3384} 
 \\ \midrule

\multirow{6}{*}{
\makecell{
FVS~\cite{wu2020future} split\\
train 256 $\times$ 832\\
test 256 $\times$ 832
}
} & 
FVS~\cite{wu2020future}* & \multicolumn{3}{c}{\na} & 0.1848 & 0.2461 & 0.3049 
\\
& SADM~\cite{bei2021learning}* & \multicolumn{3}{c}{\na} & 0.1441 & 0.2458 & 0.3116 
 \\ \cmidrule(l){2-8} 
& PredNet~\cite{lotter2016deep} & \multicolumn{3}{c}{\na} & 0.5535 & 0.5866 & 0.6295 
\\
 & MCNet~\cite{villegas2017decomposing} & \multicolumn{3}{c}{\na} & 0.2405 & \textbf{0.3171} & \textbf{0.3739} 
 \\
 & VoxelFlow~\cite{liu2017video} & \multicolumn{3}{c}{\na} & 0.3247 & 0.3743 & 0.4159
 \\
 & \model (ours) & \textbf{0.150} & \textbf{0.739} & \textbf{19.9} & \textbf{0.1560} & 0.3441 & 0.4467 
 \\ \bottomrule
\end{tabular}%
\vspace{10pt}
\caption{Results on KITTI~\cite{geiger2013vision} with three train-test splits used in previous methods. 
Our method achieves better performance than all \nvs baselines, including those pre-trained on ImageNet~\cite{deng2009imagenet} (denoted by *). We also achieve competitive performance compared with \vp baselines. 
LDI~\cite{tulsiani2018layer} and Tucker~\etal~\cite{tucker2020single} do not report LPIPS values, denoted by \na. Baselines can do only one task while ours solves both, indicated by \na. 
}
\vspace{-15pt}

\label{table:kitti}
\end{table}

We conduct extensive experiments to validate that our method compares favorably to state-of-the-art methods on both monocular \nvs and \vp in diverse scenarios: during training, our model may learn from single- and multi-view videos, static and moving cameras, and indoor and outdoor scenes. %
In particular, our method is the only one that performs both tasks simultaneously. 
Finally in \sect{section:mpis}, we provide an analysis of intermediate model outputs for an intuitive understanding of our method. %

\subsection{Learning from Multi-View Videos with Moving Cameras}

\paragraph{Setup.}
\label{section:multiple}
\label{experiments:kitti}
We first test our model on learning from videos with binocular views, as shown in~\fig{fig:setups}(a). KITTI~\cite{geiger2013vision} is a benchmark dataset widely used for both \nvs and \vp. It contains street scenes captured by two stereo cameras mounted on a moving car. 
Following prior methods~\cite{tulsiani2018layer,tucker2020single,limine}, we use the 28 city scenes from the KITTI-raw dataset and split them into 20 sequences for training, 4 for validation, and 4 for testing. 
This is denoted as LDI~\cite{tulsiani2018layer} split in \tbl{table:kitti}.
To be directly comparable with prior works on \vp~\cite{wu2020future,bei2021learning}, we also evaluate on the split with 24 sequences for training and 4 for testing, denoted as FVS~\cite{wu2020future} split.

\myparagraph{Evaluation.}
\label{evaluation}
For \nvs evaluation, images are cropped by 5\% from the boundary of all sides and resized to $128 \times 384$. We report the similarity of synthesized images compared to the ground truth using LPIPS~\cite{zhang2018unreasonable} computed with VGG features, SSIM~\cite{wang2004image} and PSNR. We report LPIPS with AlexNet features for \vp to be directly comparable to prior methods~\cite{wu2020future,bei2021learning}. 

\myparagraph{Baselines.}
For \nvs, we compare our results with
Tucker and Snavely~\cite{tucker2020single} and LDI~\cite{tulsiani2018layer}. LDI~\cite{tulsiani2018layer} derives a 2-layer representation from single-view inputs. We also compare with MINE~\cite{limine}, a recent work that extends Tucker and Snavely~\cite{tucker2020single} to the continuous depth domain using implicit functions. MINE~\cite{limine}-32 and MINE~\cite{limine}-64 from \tbl{table:kitti} refer to two model variants with 32 and 64 planes, respectively, as reported in the original paper.

For \vp, we compare our method with PredRNN-V2~\cite{wang2022predrnn}, a recently proposed method using a pair of decoupled RNN memory cells. We retrain their non-action-conditioned model on the dataset with $2$ input frames and a prediction length of $5$. 
We also compare with two \vp methods which decompose scenes for better dynamics modeling. FVS~\cite{wu2020future} predicts the next frame by decomposing frames in a video into object-centric layers and modeling the dynamics of each layer. Similar to ours, FVS also assumes the motion of each layer to be affine. Another baseline is SADM~\cite{bei2021learning} which decomposes frames into semantically consistent regions and predicts the motion of each region. Both FVS and SADM require instance maps and optical flow inputs obtained from pretrained models.

\begin{figure}[t!]
    \centering
    \includegraphics[width=\linewidth]{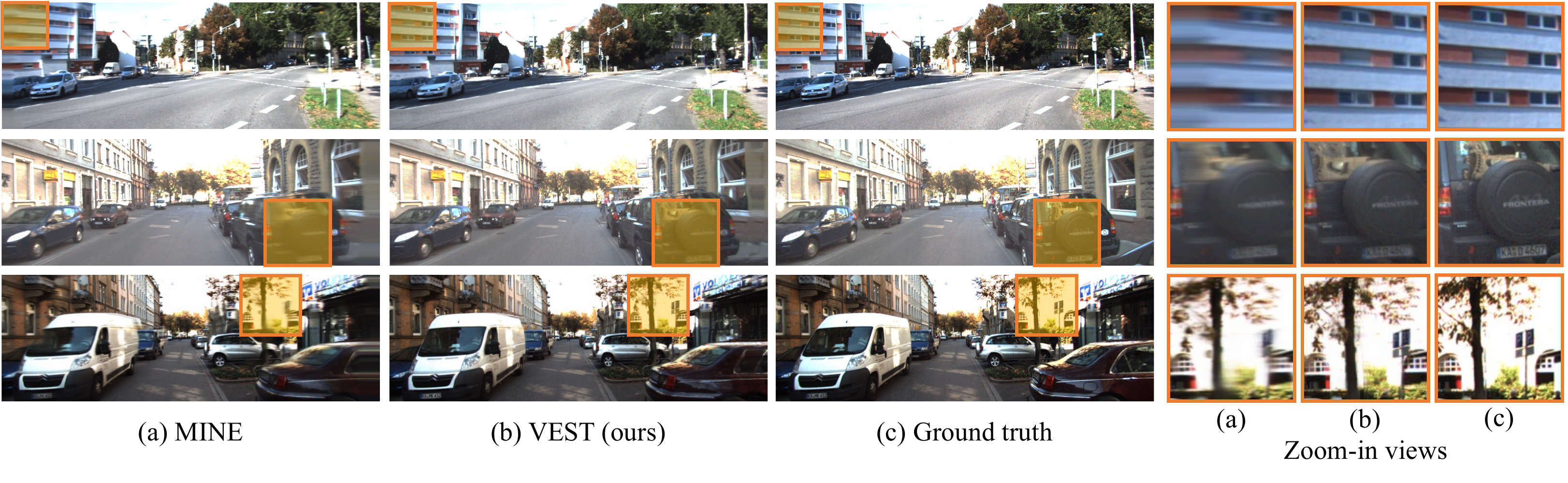}
    \vspace{-15pt}
    \caption{View synthesis results on KITTI. %
    Left: (a) results from MINE~\cite{limine}; (b) results from \model (ours); (c) ground truth images. Right: zoom-in views. In the first two examples, our model produces fine details for small objects, such as windows and printed texts. The third example shows that our model performs more accurate extrapolation even for challenging, thin objects, such as trees with small distortion.}
    \label{fig:kitti_space}
    \vspace{-15pt}
\end{figure}

\begin{figure}[t]
    \includegraphics[width=\linewidth]{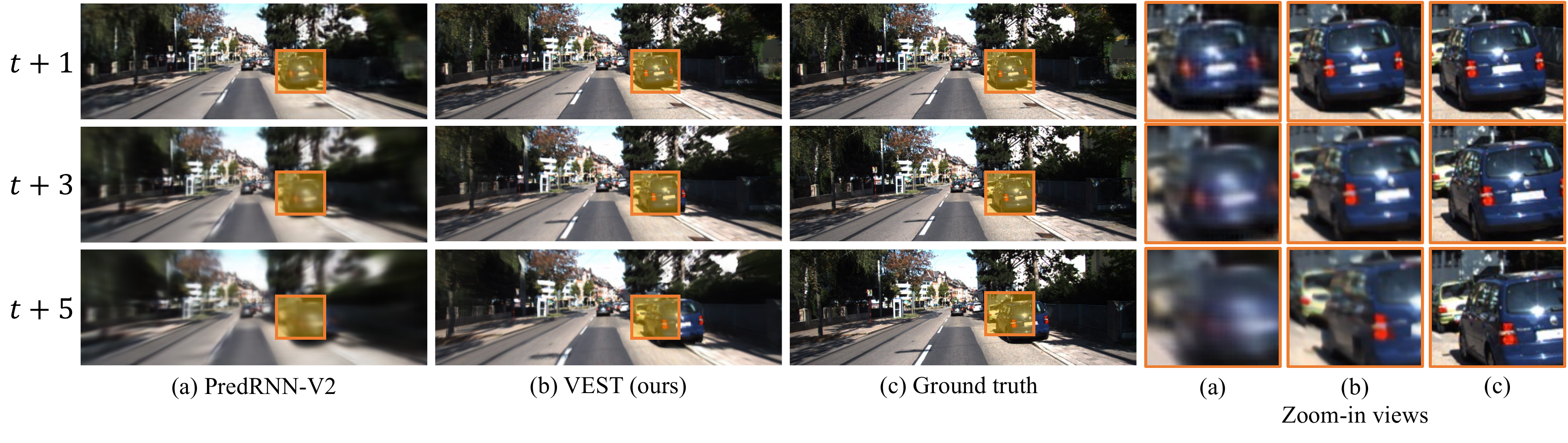}
    \vspace{-20pt}
    \caption{Video prediction results on KITTI. %
    Left: (a) predictions from PredRNN-V2~\cite{wang2022predrnn}; (b) predictions from \model (ours); (c) ground truth images. Right: zoom-in views. Each row corresponds to frame prediction for $t+1$, $t+3$, $t+5$. Our \model produces much sharper predictions compared to the baseline.
    }
    \vspace{-10pt}
    \label{fig:kitt_time}
\end{figure}

\myparagraph{Results.} Results are shown in \tbl{table:kitti}. All our models are trained with the same configuration. For \nvs, our model achieves a significant improvement across all metrics compared to the \nvs baselines including variants of MINE~\cite{limine} that are pre-trained on ImageNet~\cite{deng2009imagenet}. 
We further show a qualitative comparison in~\fig{fig:kitti_space} where MINE~\cite{limine} uses ImageNet pretraining. Please refer to the project page for better visualization. 

We additionally test our model with a different inference procedure (\model-single as shown in Table~\ref{table:kitti}). Instead of $I_{t-1}$ and $I_t$, the model receives a single source frame $I_t$ repeated twice as inputs to predict $I_t^\prime$. In this way, our model receives no additional information from historical frames. \model and \model-single from Table~\ref{table:kitti} use the same checkpoint and both outperform all non-pretrained baselines across all metrics. 
 
For \vp, our approach leads to a significant improvement compared to PredRNN-V2~\cite{wang2022predrnn}. \fig{fig:kitt_time} shows that our model makes prediction with sharper edges and higher fidelity. 
In these examples, street and cars are moving towards the camera, which is effectively a scale transformation in 2D. Because scale transformation is a subclass of affine transformation, our method can accurately capture such perpendicular motions in the scenes.
All qualitative results from the KITTI dataset are used under license CC BY-NC-SA 3.0. 

\myparagraph{Ablation.} We ablate the importance of using Perceptual loss for \nvs. In Table~\ref{table:kitti}, \model-no-perc corresponds to a variant with $\lambda_\text{perc}^\text{space} = 0$, suggesting that perceptual loss in the spatial loss term largely improves performance. 
We also include an ablation study on the effect of the number of MPI planes in \sect{sec:ablate_num_planes}.

\subsection{Learning from Single-View Videos with a Moving Camera}
\label{section:single}

\paragraph{Setup.} 
We evaluate our method on video data with a single view, as shown in~\fig{fig:setups}(b). RealEstate10K~\cite{zhou2018stereo} is a standard \nvs benchmark dataset consisting of 80K videos filming mostly indoor scenes. We follow the training-test split from Lai~\etal~\cite{lai2021video}, which uses a randomly sampled subset of the full dataset. It includes 10K video sequences for training and 5K for testing. 
The training and evaluation resolution is $256 \times 256$. 

\myparagraph{Evaluation.}
Following MINE~\cite{limine}, for evaluation, we randomly sample 5 source frames from each testing sequence and sample target frames that are 5, 10, or at most 30 frames apart for each of the source frames. 
We evaluate on the intersection of the testing frames used in MINE and our testing split. This results in 372 testing sequences in total. The similarity scores are measured with LPIPS~\cite{zhang2018unreasonable} with VGG features, SSIM~\cite{wang2004image}, and PSNR. 

\myparagraph{Baselines.}
We run the publicly released checkpoint from MINE~\cite{limine} with 64 planes, which is pre-trained on ImageNet and trained on the full training split of RealEstate10K with a resolution of $384 \times 256$. We first run their model with resolution $384 \times 256$, and then downsample to $256 \times 256$ to compare with the ground truth. Both our model and MINE receive the same set of sparse points for scale-invariant analysis in \eqn{depth_scale}. We also compare with Synsin~\cite{wiles2020synsin} and Tucker and Snavely~\cite{tucker2020single}, both taking single-view inputs, and Zhou~\etal~\cite{zhou2018stereo}, which takes in binocular inputs.

\begin{table}[t]
\centering\scriptsize
\setlength{\tabcolsep}{2pt}

\begin{tabular}{lcccccccccccc}
\toprule
 \multirow{2}{*}{Method} & \multicolumn{2}{c}{LPIPS$\downarrow$} &
 \multicolumn{2}{c}{SSIM$\uparrow$} & \multicolumn{2}{c}{PSNR$\uparrow$} 
 &LPIPS&SSIM&PSNR\\ \cmidrule(l){2-3} \cmidrule(l){4-5} \cmidrule(l){6-7} \cmidrule(l){8-10}
& $n=5$  & $n=$  rand & $n=5$ & $n=$ rand & $n=5$   & $n=$  rand & \multicolumn{3}{c}{$t+1$} \\ \midrule
MINE~\cite{limine}* & 0.0986  & 0.1774 & 0.9018 & 0.8221 &  27.9837  & 24.3112 & \multicolumn{3}{c}{\na} \\ \midrule
SynSin~\cite{wiles2020synsin} & \na&\na&\na &   0.7400 & \na &  22.3100 &  \multicolumn{3}{c}{\na}\\
Tucker~\etal~\cite{tucker2020single} & 0.0967 & 0.1761 & 0.8699 & 0.7851 & 27.0500  & 23.5200 &  \multicolumn{3}{c}{\na}\\
\model-single & 0.0944 & 0.1736 &  0.8700 & 0.7688 &  26.6599 & 24.6906 &  \multicolumn{3}{c}{\na} \\
Zhou~\etal~\cite{zhou2018stereo} & \textbf{0.0816}  & 0.1667 &  0.8943 & \textbf{0.8014}  & 27.5788& 24.1531 &  \multicolumn{3}{c}{\na}\\ 
\model (ours) & 0.0841 & \textbf{0.1596} & \textbf{0.8987} & 0.8003 & \textbf{28.2078} & \textbf{25.7553} & 0.0436  & 0.9334& 31.6831 \\ \bottomrule
\end{tabular}%

\vspace{5pt}
\caption{Results on RealEstate10K~\cite{zhou2018stereo}. Our model achieves a better or comparable performance compared to the baselines, and only ours can predict future frames. SynSin~\cite{wiles2020synsin} does not report LPIPS values, denoted by \na. $n$ specifies the number of frames between source and target images in the video sequence, and $n=$ rand means that $n$ is uniformly sampled between $1$ and $30$. A larger $n$ indicates larger range of spatial extrapolation. * indicates that the model requires pretraining. 
}
\label{table:estate}
\vspace{-10pt}
\end{table}
\begin{figure}[t!]
    \centering
    \includegraphics[width=\linewidth]{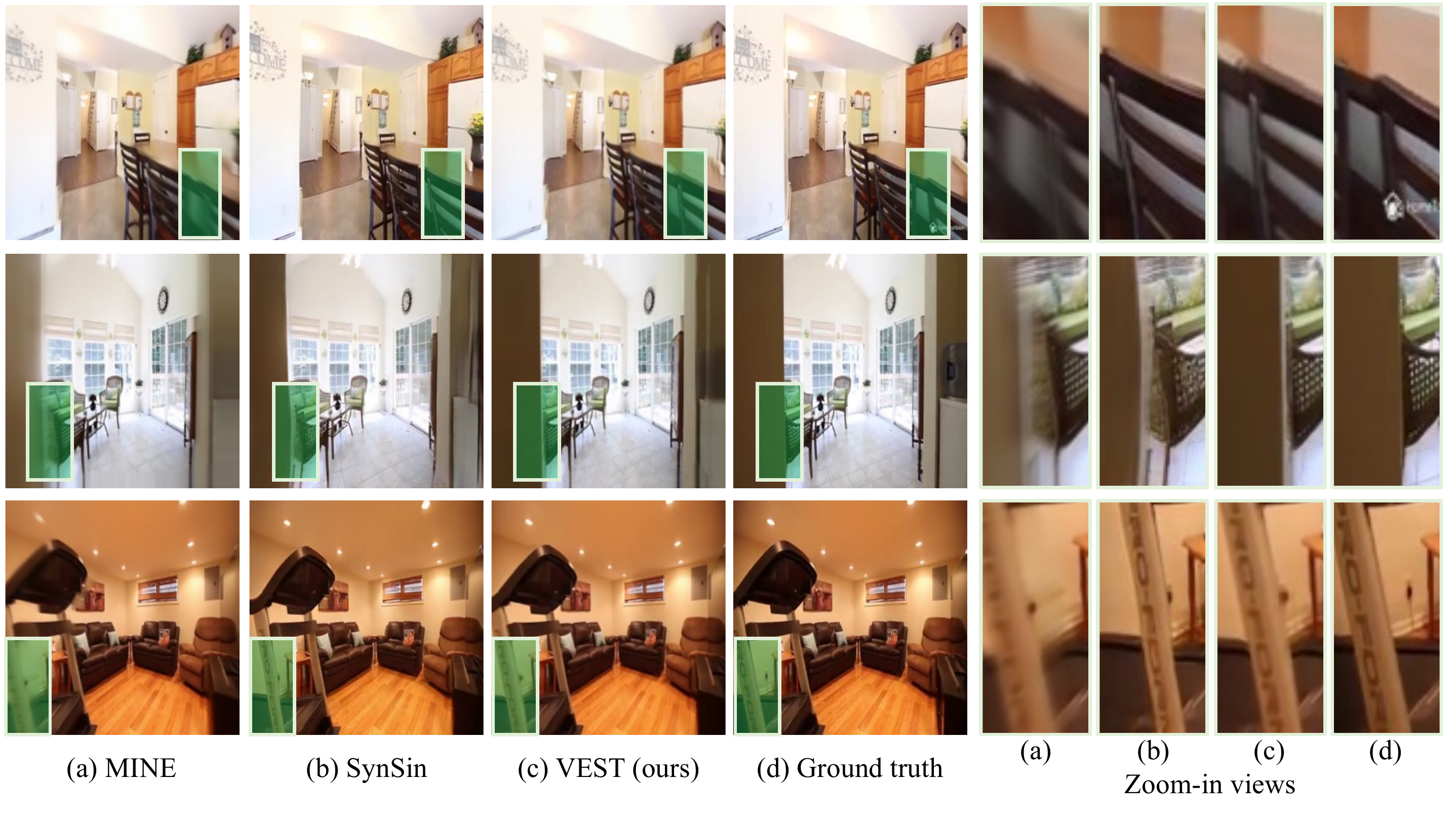}
    \vspace{-15pt}
    \caption{Novel view synthesis results on RealEstate10K. 
    (a)-(c) correspond to results from MINE~\cite{limine}, SynSin~\cite{wiles2020synsin}, and \model (ours). (d) corresponds to the ground truth target image $I_{22}$. These examples show that \model produces sharper details than MINE~\cite{limine}. It also predicts object positions more accurately compared to SynSin~\cite{wiles2020synsin}. }
    \vspace{-10pt}
    \label{fig:estate_space}
\end{figure}

\begin{figure}[t!]
    \centering
    \includegraphics[width=\linewidth]{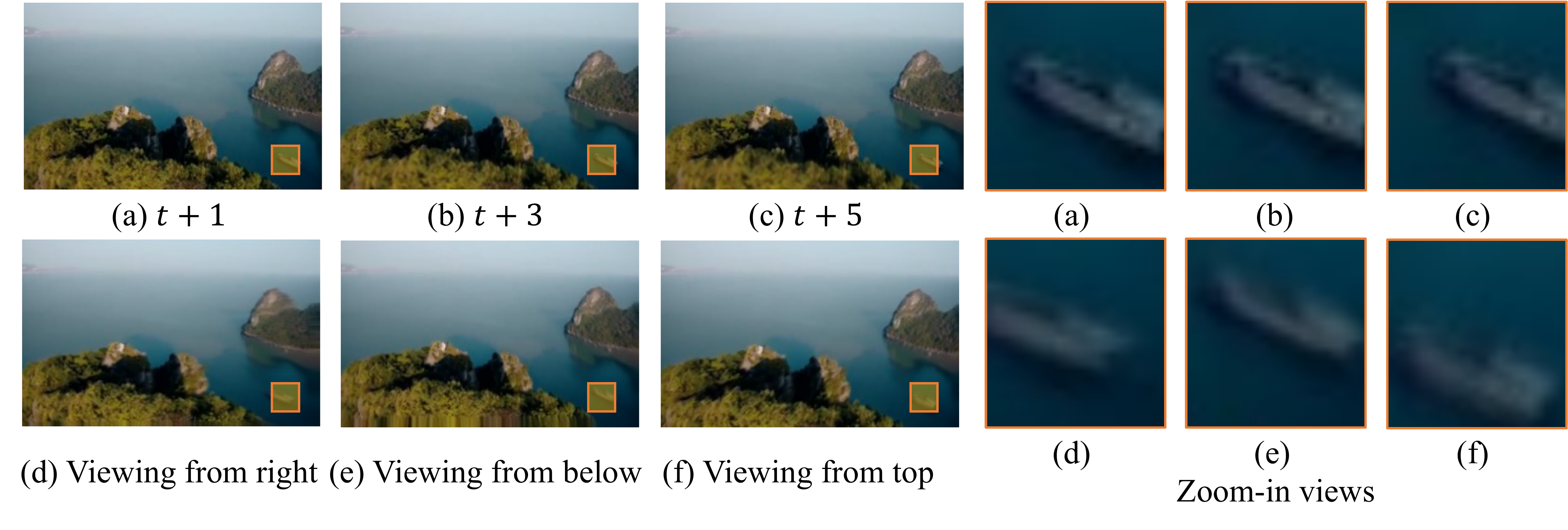}
    \vspace{-20pt}
    \caption{Results on the ACID dataset filmed by a monocular moving camera. (a)-(c) are \vp results and (d)-(f) \nvs results. 
    }
    \label{fig:acid}
\end{figure}

\myparagraph{Results.}
We show quantitative results in Table~\ref{table:estate}.
We obtain similar scores as reported in the original paper.\footnote{The LPIPS scores of MINE~\cite{limine} computed are slightly worse compared to the original paper due to a bug in the evaluation script in their public codebase, where tensors in range $[0, 1]$ are fed into an LPIPS package which expects inputs in range $[-1, 1]$.}
Results for Tucker and Snavely~\cite{tucker2020single} and SynSin~\cite{wiles2020synsin} are taken from the original paper for reference, and they are evaluated on a different test split. Our method outperforms MINE~\cite{limine}, a state-of-the-art single-view \nvs method on this dataset, despite that the baseline uses ImageNet pre-training and trains on the full dataset. We show qualitative comparisons in~\fig{fig:estate_space}.

On this dataset, modeling the motion is beneficial as our method uses the motion parallax from input frames by modeling the dynamics as opposed to explicit plane sweep volume construction as done in the stereo-based baselines ~\cite{zhou2018stereo,lin2021deep}. Indeed, our method is comparable to the binocular baseline Zhou~\etal~\cite{zhou2018stereo}. 

Additionally, we train our model on Aerial Coastline Imagery Dataset (ACID)~\cite{liu2021infinite}, a single-view dataset mostly filming outdoor natural scenes. 
Our method can synthesize images corresponding to the queried viewpoint and timestamp, as shown in \fig{fig:acid}. Both RealEstate10K and ACID datasets include YouTube videos under the Creative-Commons license.

\begin{table}[t]
\centering\scriptsize
\begin{tabular}{clccc}
\toprule
Test-time Input & Method & LPIPS$\downarrow$ & \multicolumn{1}{l}{SSIM$\uparrow$} & PSNR$\uparrow$ \\ 
\midrule 
Multi-view &Lin~\etal~\cite{lin2021deep} & 0.1558 & 0.8667 & 21.1988 \\ 
\midrule
\multirow{3}{*}{Single-view} &Lin~\etal~\cite{lin2021deep} & 0.3719 & 0.4929 & 16.7794 \\ 
 & \model-single (ours) & 0.2600 &0.6242 & 21.6957 \\
 & \model (ours) & {\bf 0.2591} & {\bf 0.6249} & \textbf{21.7664} \\ 
\bottomrule
\end{tabular}%
\vspace{5pt}
\caption{Results on the dataset from Lin~\etal~\cite{lin2021deep} filmed by multi-view static cameras. Our method achieves better performance than the baseline method under the same view setting.}
\vspace{-25pt}
\label{table:deep3d}
\end{table}
\begin{figure}[t]
    \centering%
    \includegraphics[width=\linewidth]{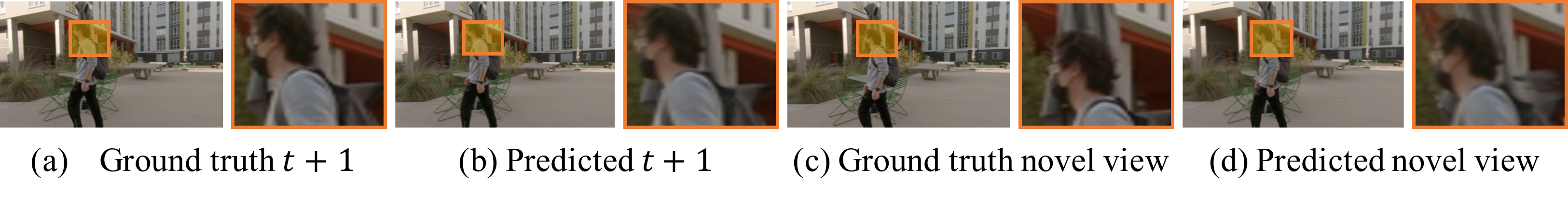}
    \vspace{-20pt}
    \caption{Results on dataset from Lin~\etal~\cite{lin2021deep}.
    (a)-(b) \vp results; (c)-(d) \nvs results. 
    Our method assigns the moving human figure and the background scene into different MPI layers. 
    }
    \label{fig:deep3d_layers}
    \vspace{-5pt}
\end{figure}

\subsection{Learning from Multi-View Videos with Static Cameras}
\label{section:multi_static}

\paragraph{Setup.} We evaluate our method on learning from a multi-view dataset with static cameras. We use the dataset from Lin~\etal~\cite{lin2021deep}, containing videos captured by 10 synchronized, static cameras, where scenes have a static background and human body movement in the foreground. The dataset is split into 86 scenes for training and 10 scenes for testing. The training and evaluation resolution is 256 $\times$ 256. We follow Lin~\etal~\cite{lin2021deep} and pretrain our model on RealEstate10K.

\myparagraph{Results.} 
We compare with Lin~\etal~\cite{lin2021deep}, a state-of-the-art multi-view \nvs method, which incorporates a learned 3D mask volume into the MPI representation to improve the temporal consistency of MPI planes. 

As shown in \tbl{table:deep3d}, our model outperforms the baseline on novel view synthesis from a single view. Note that Lin~\etal~\cite{lin2021deep} was originally designed to perform view synthesis from stereo inputs and works well in their setup; in contrast, our model is designed to take single-view input during inference. We further show qualitative results in \fig{fig:deep3d_layers}. 

\begin{figure}[t]
    \centering%
    \includegraphics[width=\linewidth]{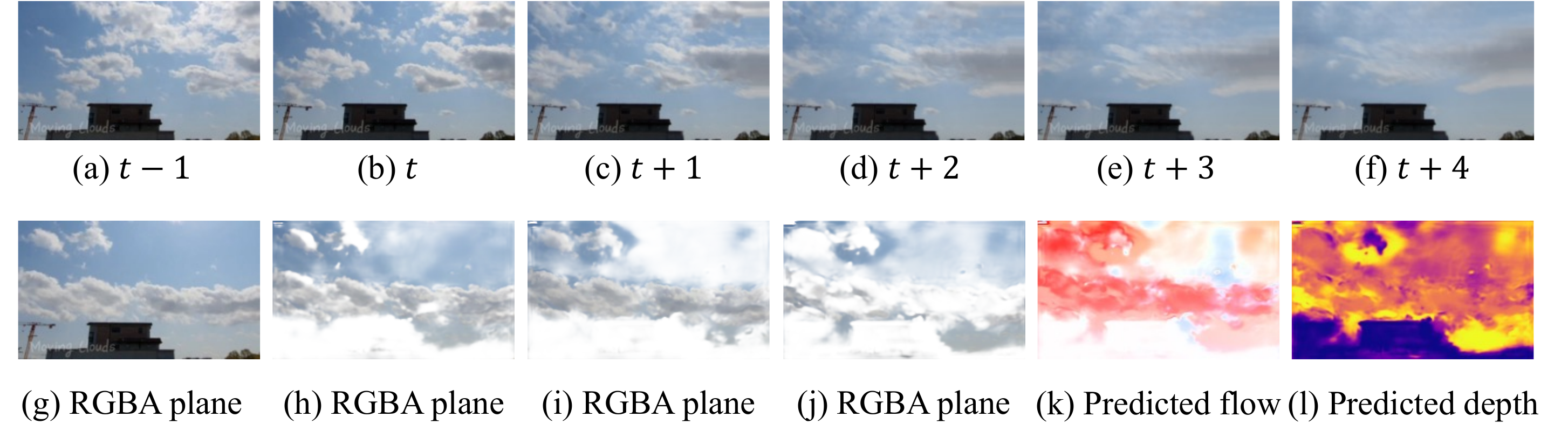}
    \vspace{-15pt}
    \caption{Video prediction results on the cloud dataset. (a)-(b) are two input frames, and (c)-(f) are predicted future frames. (g)-(j) are RGBA planes ordered from far to near, and (k)-(l) are the flow and depth maps predicted by the model. 
    In this example, regions of clouds and the building are assigned to different planes, corresponding to different motions.}
    \label{fig:cloud}

    \vspace{-15pt}
\end{figure}

\subsection{Learning from Single-View Videos with a Static Camera}
\label{section:single_static}

In the last camera setting, videos are filmed by a single, static camera, and \nvs task becomes non-applicable due to a lack of viewpoint changes throughout a video sequence. 
While our method is motivated by leveraging the spatial and temporal cues from scenes, we evaluate our method under this camera setting for completeness, and show that our method can decompose scenes into layers based on different motion patterns. 

We collect a dataset of 175 moving clouds videos from YouTube, which will be made publicly available. 
Our method produces realistic \vp results as shown in \fig{fig:cloud}. Since each affine transformation layer is weighted by the predicted alpha map, the overall scene dynamics is not restricted to be affine. 
We defer further analysis on modeling scene dynamics in such camera setting to \sectapp{appendix:dynamics}.

\begin{figure}[t!]
    \centering
    \includegraphics[width=\linewidth]{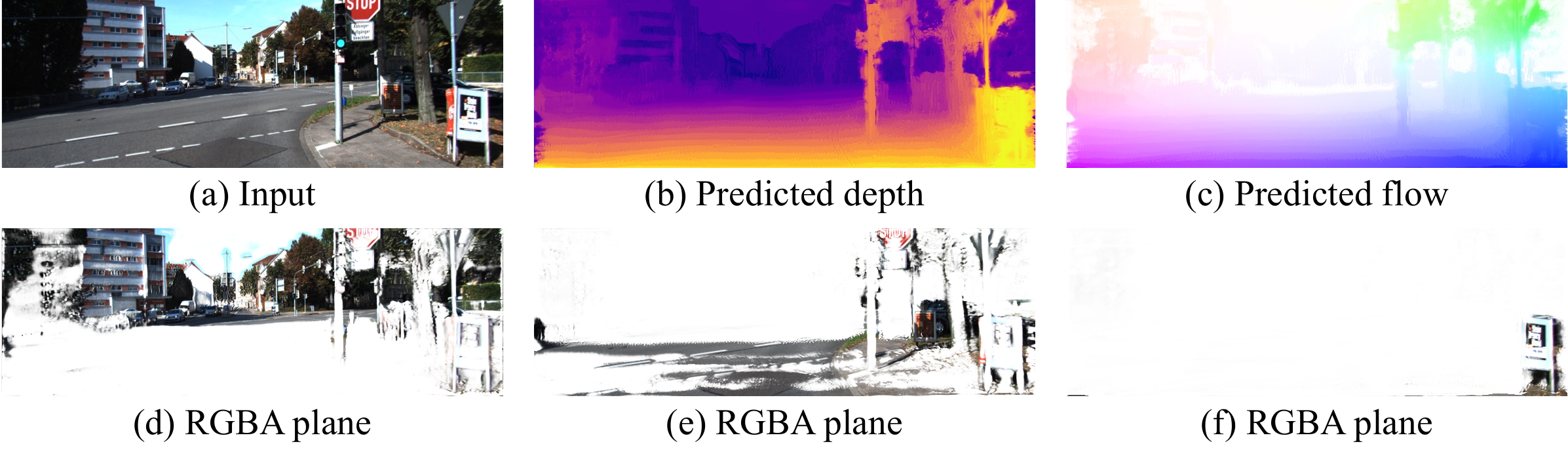}
    \vspace{-15pt}
    \caption{Visualization of intermediate predictions on KITTI dataset. (a) The second input frame, which is the source image for MPI representations. (b) and (c) are depth and flow predictions. In (d-f), we show 3 out of 16 RGBA planes predicted by the model, ordered from far to near. }
    \label{fig:maps}
    \vspace{-10pt}
\end{figure}

\subsection{Qualitative Analysis of the Generalized MPIs}
\label{section:mpis}
To understand how the model learns to solve \nvs and \vp simultaneously, we qualitatively analyze the intermediate model outputs 
as shown in \fig{fig:maps}. 
The depth map is computed from \eqn{depth_composition}, and the flow map is composed from per-layer flow maps similarly. 
In the shown example, the camera is moving forward and objects closer to the camera tend to have more dominant motions. 
With the generalized MPI representation, the scene dynamics is decomposed into plane-wise motion fields. Since each plane is depth-aware, this representation helps introduce an inductive bias for the model to learn the depth-motion correspondence which facilitates the learning of the two extrapolation tasks.

\section{Conclusion}
In this work, we view \nvs and \vp as extrapolation along two axes for the spatial-temporal coordinates of videos. 
\nvs utilizes camera viewpoint changes in a video sequence to discover depth, while \vp considers both camera and object motions. The two tasks can be jointly learned to develop a scene representation from video data with complementary learning signals coming from each of the tasks. We propose a generalized MPI representation to tackle both tasks, and develop a model that achieves superior or comparable performance compared to previous methods that tackle only one of the tasks, on natural datasets for indoor and outdoor scenes. Please see \sectapp{appendix:discussions} for more discussions.

\myparagraph{Acknowledgement.}
We thank Angjoo Kanazawa, Hong-Xing (Koven) Yu, Huazhe (Harry) Xu, Noah Snavely, Ruohan Zhang, Ruohan Gao, and Shangzhe (Elliott) Wu for detailed feedback on the paper, and Kaidi Cao for collecting the cloud dataset. This work is in part supported by the Stanford Institute for Human-Centered AI (HAI), the Stanford Center for Integrated Facility Engineering (CIFE), the Samsung Global Research Outreach (GRO) Program, and Amazon, Autodesk, Meta, Google, Bosch, and Adobe.

\bibliographystyle{splncs04}
\bibliography{main}

\begin{thebibliography}{10}
\providecommand{\url}[1]{\texttt{#1}}
\providecommand{\urlprefix}{URL }
\providecommand{\doi}[1]{https://doi.org/#1}

\bibitem{bei2021learning}
Bei, X., Yang, Y., Soatto, S.: Learning semantic-aware dynamics for video
  prediction. In: CVPR (2021)

\bibitem{deng2009imagenet}
Deng, J., Dong, W., Socher, R., Li, L.J., Li, K., Fei-Fei, L.: {ImageNet}: A
  large-scale hierarchical image database. In: CVPR (2009)

\bibitem{du2021neural}
Du, Y., Zhang, Y., Yu, H.X., Tenenbaum, J.B., Wu, J.: Neural radiance flow for
  {4D} view synthesis and video processing. In: ICCV (2021)

\bibitem{flynn2019deepview}
Flynn, J., Broxton, M., Debevec, P., DuVall, M., Fyffe, G., Overbeck, R.,
  Snavely, N., Tucker, R.: {DeepView}: View synthesis with learned gradient
  descent. In: CVPR (2019)

\bibitem{gao2019disentangling}
Gao, H., Xu, H., Cai, Q.Z., Wang, R., Yu, F., Darrell, T.: Disentangling
  propagation and generation for video prediction. In: ICCV (2019)

\bibitem{geiger2013vision}
Geiger, A., Lenz, P., Stiller, C., Urtasun, R.: Vision meets robotics: The
  {KITTI} dataset. The International Journal of Robotics Research
  \textbf{32}(11),  1231--1237 (2013)

\bibitem{girdhar2020cater}
Girdhar, R., Ramanan, D.: {CATER}: A diagnostic dataset for compositional
  actions and temporal reasoning. In: ICLR (2020)

\bibitem{hu2021worldsheet}
Hu, R., Ravi, N., Berg, A.C., Pathak, D.: Worldsheet: Wrapping the world in a
  3d sheet for view synthesis from a single image. In: ICCV (2021)

\bibitem{kingma2015adam}
Kingma, D.P., Ba, J.: Adam: A method for stochastic optimization. In: ICLR
  (2015)

\bibitem{lai2021video}
Lai, Z., Liu, S., Efros, A.A., Wang, X.: Video autoencoder: self-supervised
  disentanglement of static {3D} structure and motion. In: ICCV (2021)

\bibitem{limine}
Li, J., Feng, Z., She, Q., Ding, H., Wang, C., Lee, G.H.: {MINE}: Towards
  continuous depth mpi with nerf for novel view synthesis. In: ICCV (2021)

\bibitem{li2021neural}
Li, Z., Niklaus, S., Snavely, N., Wang, O.: Neural scene flow fields for
  space-time view synthesis of dynamic scenes. In: CVPR (2021)

\bibitem{lin2021deep}
Lin, K.E., Xiao, L., Liu, F., Yang, G., Ramamoorthi, R.: Deep {3D} mask volume
  for view synthesis of dynamic scenes. In: ICCV (2021)

\bibitem{liu2021infinite}
Liu, A., Tucker, R., Jampani, V., Makadia, A., Snavely, N., Kanazawa, A.:
  Infinite nature: Perpetual view generation of natural scenes from a single
  image. In: ICCV (2021)

\bibitem{liu2017video}
Liu, Z., Yeh, R., Tang, X., Liu, Y., Agarwala, A.: Video frame synthesis using
  deep voxel flow. In: ICCV (2017)

\bibitem{lotter2016deep}
Lotter, W., Kreiman, G., Cox, D.: Deep predictive coding networks for video
  prediction and unsupervised learning. In: ICLR (2017)

\bibitem{lu2020}
Lu, E., Cole, F., Dekel, T., Xie, W., Zisserman, A., Salesin, D., Freeman,
  W.T., Rubinstein, M.: Layered neural rendering for retiming people in video.
  In: SIGGRAPH Asia (2020)

\bibitem{lu2021omnimatte}
Lu, E., Cole, F., Dekel, T., Zisserman, A., Freeman, W.T., Rubinstein, M.:
  Omnimatte: Associating objects and their effects in video. In: CVPR (2021)

\bibitem{mildenhall2020nerf}
Mildenhall, B., Srinivasan, P.P., Tancik, M., Barron, J.T., Ramamoorthi, R.,
  Ng, R.: {NeRF}: Representing scenes as neural radiance fields for view
  synthesis. In: ECCV (2020)

\bibitem{park2021nerfies}
Park, K., Sinha, U., Barron, J.T., Bouaziz, S., Goldman, D.B., Seitz, S.M.,
  Martin-Brualla, R.: Nerfies: Deformable neural radiance fields. In: ICCV
  (2021)

\bibitem{pumarola2021d}
Pumarola, A., Corona, E., Pons-Moll, G., Moreno-Noguer, F.: D-nerf: Neural
  radiance fields for dynamic scenes. In: CVPR (2021)

\bibitem{ranzato2014video}
Ranzato, M., Szlam, A., Bruna, J., Mathieu, M., Collobert, R., Chopra, S.:
  Video (language) modeling: a baseline for generative models of natural
  videos. arXiv preprint arXiv:1412.6604  (2014)

\bibitem{schoenberger2016sfm}
Sch\"{o}nberger, J.L., Frahm, J.M.: Structure-from-motion revisited. In: CVPR
  (2016)

\bibitem{schoenberger2016mvs}
Sch\"{o}nberger, J.L., Zheng, E., Pollefeys, M., Frahm, J.M.: Pixelwise view
  selection for unstructured multi-view stereo. In: ECCV (2016)

\bibitem{shade1998layered}
Shade, J., Gortler, S., He, L.w., Szeliski, R.: Layered depth images. In:
  SIGGRAPH (1998)

\bibitem{shi2015convolutional}
Shi, X., Chen, Z., Wang, H., Yeung, D.Y., Wong, W.K., Woo, W.c.: Convolutional
  {LSTM} network: A machine learning approach for precipitation nowcasting. In:
  NeurIPS (2015)

\bibitem{shih20203d}
Shih, M.L., Su, S.Y., Kopf, J., Huang, J.B.: {3D} photography using
  context-aware layered depth inpainting. In: CVPR (2020)

\bibitem{simonyan2014very}
Simonyan, K., Zisserman, A.: Very deep convolutional networks for large-scale
  image recognition. In: ICLR (2015)

\bibitem{srinivasan2019pushing}
Srinivasan, P.P., Tucker, R., Barron, J.T., Ramamoorthi, R., Ng, R., Snavely,
  N.: Pushing the boundaries of view extrapolation with multiplane images. In:
  CVPR (2019)

\bibitem{srivastava2015unsupervised}
Srivastava, N., Mansimov, E., Salakhutdinov, R.: Unsupervised learning of video
  representations using {LSTM}s. In: ICML (2015)

\bibitem{tretschk2021non}
Tretschk, E., Tewari, A., Golyanik, V., Zollh{\"o}fer, M., Lassner, C.,
  Theobalt, C.: Non-rigid neural radiance fields: Reconstruction and novel view
  synthesis of a dynamic scene from monocular video. In: ICCV (2021)

\bibitem{tucker2020single}
Tucker, R., Snavely, N.: Single-view view synthesis with multiplane images. In:
  CVPR. pp. 551--560 (2020)

\bibitem{tulsiani2018layer}
Tulsiani, S., Tucker, R., Snavely, N.: Layer-structured {3D} scene inference
  via view synthesis. In: ECCV (2018)

\bibitem{villegas2017decomposing}
Villegas, R., Yang, J., Hong, S., Lin, X., Lee, H.: Decomposing motion and
  content for natural video sequence prediction. In: ICLR (2017)

\bibitem{wang1993layered}
Wang, J.Y.A., Adelson, E.H.: Layered representation for motion analysis. In:
  CVPR (1993)

\bibitem{wang2022predrnn}
Wang, Y., Wu, H., Zhang, J., Gao, Z., Wang, J., Yu, P., Long, M.: Predrnn: A
  recurrent neural network for spatiotemporal predictive learning. IEEE TPAMI
  (2022)

\bibitem{wang2004image}
Wang, Z., Bovik, A.C., Sheikh, H.R., Simoncelli, E.P.: Image quality
  assessment: From error visibility to structural similarity. IEEE TIP
  \textbf{13}(4),  600--612 (2004)

\bibitem{wiles2020synsin}
Wiles, O., Gkioxari, G., Szeliski, R., Johnson, J.: {SynSin}: End-to-end view
  synthesis from a single image. In: CVPR (2020)

\bibitem{wu2020future}
Wu, Y., Gao, R., Park, J., Chen, Q.: Future video synthesis with object motion
  prediction. In: CVPR (2020)

\bibitem{xian2021space}
Xian, W., Huang, J.B., Kopf, J., Kim, C.: Space-time neural irradiance fields
  for free-viewpoint video. In: CVPR (2021)

\bibitem{yoon2020novel}
Yoon, J.S., Kim, K., Gallo, O., Park, H.S., Kautz, J.: Novel view synthesis of
  dynamic scenes with globally coherent depths from a monocular camera. In:
  CVPR. pp. 5336--5345 (2020)

\bibitem{yu2021pixelnerf}
Yu, A., Ye, V., Tancik, M., Kanazawa, A.: {pixelNeRF}: Neural radiance fields
  from one or few images. In: CVPR (2021)

\bibitem{zhang2018unreasonable}
Zhang, R., Isola, P., Efros, A.A., Shechtman, E., Wang, O.: The unreasonable
  effectiveness of deep networks as a perceptual metric. In: CVPR (2018)

\bibitem{zhou2018stereo}
Zhou, T., Tucker, R., Flynn, J., Fyffe, G., Snavely, N.: Stereo magnification:
  Learning view synthesis using multiplane images. In: SIGGRAPH (2018)

\bibitem{zhou2016view}
Zhou, T., Tulsiani, S., Sun, W., Malik, J., Efros, A.A.: View synthesis by
  appearance flow. In: ECCV (2016)

\end{thebibliography}

\clearpage
\appendix

\section{Architecture Details}
\label{appendix:architecture}
The architecture used for the MPI encoder is specified in Table~\ref{table:architecture}.

\begin{table}[h]
\centering\footnotesize
\begin{tabular}{@{}cccccccc@{}}
\toprule
\textbf{Input} & \textbf{k} & \textbf{c} & \textbf{Output} & \textbf{Input} & \textbf{k} & \textbf{c} & \textbf{Output} \\ \midrule
Concat($I_{t-1}, I_t$) & 7 & 32 & down1 & down1 & 7 & 32 & down1b \\
MP2(down1b) & 5 & 64 & down2 & down2 & 5 & 64 & down2b \\
MP2(down2b) & 3 & 128 & down3 & down3 & 3 & 128 & down3b \\
MP2(down3b) & 3 & 256 & down4 & down4 & 3 & 256 & down4b \\
MP2(down4b) & 3 & 512 & down5 & down5 & 3 & 512 & down5b \\
MP2(down5b) & 3 & 512 & down6 & down6 & 3 & 512 & down6b \\
MP2(down6b) & 3 & 512 & mid1 & mid1 & 3 & 512 & mid2 \\
Up2(mid2) + down6b & 3 & 512 & up6 & up6 & 3 & 512 & up6b \\
Up2(up6b) + down5b & 3 & 512 & up5 & up5 & 3 & 512 & up5b \\
Up2(up5b) + down4b & 3 & 256 & up4 & up4 & 3 & 256 & up4b \\
Up2(up4b) + down3b & 3 & 128 & up3 & up3 & 3 & 128 & up3b \\
Up2(up3b) + down2b & 3 & 64 & up2 & up2 & 3 & 64 & up2b \\
Up2(up2b) + down1b & 3 & 64 & post1 & post1 & 3 & 64 & post2 \\
post2 & 3 & 64 & up1 & up1 & 3 & 64 & up1b \\
up1b & 3 & 64 x D & conv1 & Reshape(conv1) & 3 & 64 & conv2 \\
conv2 & 7 & 7 & conv3 & ReshapeBack(conv3) & - & - & output \\ \bottomrule
\end{tabular}
\vspace{10pt}
\caption{MP2 is max pooling with stride 2, Up2 is nearest-neighbor upsampling with scale 2, + is concatenation. Reshape transforms a tensor with $C \times D$ channels into $C$ channels, and $D$ is merged to the batch dimension, and ReshapeBack is the reverse operation. All layers up till up1b use ReLU activation and the layers for conv1, conv2 and conv3 use LeakyReLu with a negative slope $0.2$. There is no activation following the very last layer. All layers use Instance Norm for activation normalization and Spectral Norm for weight normalization. 
}
\label{table:architecture}
\end{table}

\section{Implementation details}

To have a better gradient flow, similar to Tucker~\etal~\cite{tucker2020single}, we add a harmonious bias $1/i$ to the alpha channel prediction, so that $w_i$ from Equation (12) becomes uniformly $1/D$ during initialization. We also add an identity bias to $f^\theta$ such that each MPI plane is associated with zero motion during initialization.

In all experiments, we set the number of MPI planes to be $D = 16$. The depth values for MPI planes are linear in the inverse space, with $d_1 = 1000$ and $d_D = 1$. 

\section{Training details}
\subsection{KITTI}

Since videos from KITTI are taken by stereo cameras with fixed relative poses, the depth scale is consistent across scenes and therefore we set it to be a constant $\sigma = 1$. We use $\lambda_{1}^\text{space} = 1000$, $\lambda_\text{spec}^\text{space} = 100$, $\lambda_{1}^\text{time} = 1000$, and $ \lambda_\text{perc}^\text{time} = 10$. 
We use Adam Optimizer~\cite{kingma2015adam} with an initial learning rate $0.0002$, which we exponentially decrease by a factor of $0.8$ for every 5 epochs. We train our model for 200K iterations on two NVIDIA TITAN RTX GPUs for about two days. During training, we apply horizontal flip with 50\% probability and apply color jittering as data augmentation.

\subsection{RealEstate10K}
We train our model for 200K iterations on one NVIDIA GeForce RTX 3090 GPU, which takes about one day. We use $\mathcal{L}_1^\text{space} = 10, \mathcal{L}_\text{perc}^\text{space} = 10, \mathcal{L}_1^\text{time} = 10, \mathcal{L}_\text{perc}^\text{time} = 0$. We use Adam Optimizer~\cite{kingma2015adam} with a constant learning rate $0.0002$. 

\begin{table}[t]
    \centering\footnotesize
    \begin{tabular}{ccccccc}
    \toprule
    \multicolumn{1}{l}{\multirow{2}{*}{$D$}} & \multicolumn{3}{c}{Extrapolation in Space} & \multicolumn{3}{c}{Extrapolation in Time} \\\cmidrule(l){2-4} \cmidrule(l){5-7} 
     \multicolumn{1}{l}{} & LPIPS$\downarrow$ & PSNR $\uparrow$& SSIM $\uparrow$& LPIPS$\downarrow$ & PSNR$\uparrow$ & SSIM$\uparrow$ \\ \midrule 
     4 & 0.0987 & 19.3453 & 0.7180 & 0.0792 & 22.9415 & 0.7880 \\
     8 & 0.0874 & 20.5795 & 0.7881 & 0.0784 & 23.1073 & 0.7922 \\
     16 & 0.0786 & 21.1889 & 0.8188 & 0.0757 & 23.3812 & 0.7971 \\
     32 & 0.0762 & 21.2279 & 0.8207 & 0.0726 & 23.7882 & 0.8083 \\ 
     \bottomrule
    \end{tabular}%
    \vspace{10pt}
    \caption{Ablation on the number of MPI planes $D$. Increasing the plane count improves the performance but also increases the training time. We adopt $D=16$ in the main paper since further increasing $D$ results in diminishing returns.}
    \label{table:ablate}
    \end{table}
    
\subsection{Ablations on the number of MPI planes}
\label{sec:ablate_num_planes}
To study the effect of the number of MPI planes, we perform an ablation study on the KITTI~\cite{geiger2013vision} dataset with resolution $128\times 384$. 
As shown in \tbl{table:ablate},
a small number of MPI planes ($D=4$ or 8) results in degraded model performance. Further increasing the number of planes from 16 to 32 results in marginal performance gain, with a cost of $2.1\times$ slower training time. Therefore, we use $D=16$ for all other experiments. 
\begin{table}[t]
\centering\footnotesize
\begin{tabular}{lccc}
\toprule
Method &  \multicolumn{1}{l}{LPIPS$\downarrow$} & \multicolumn{1}{l}{PSNR$\uparrow$} & \multicolumn{1}{l}{SSIM$\uparrow$} \\ \midrule 

 PredRNN~\cite{wang2022predrnn} &0.0600  &  37.02 & 0.9643 \\
  Ours &\textbf{0.0122} & \textbf{42.58}  &\textbf{0.9762} \\  \bottomrule

\end{tabular}%
\vspace{10pt}
\caption{Results of next-frame prediction on CATER~\cite{girdhar2020cater}. Our model achieves better performance compared to PredRNN~\cite{wang2022predrnn}.}
\label{table:cater}
\end{table}

\subsection{Modeling dynamic scenes}
\label{appendix:dynamics}
To test whether our method is able to model more dynamic scenes, we test our method on CATER~\cite{girdhar2020cater}, a dataset of scenes with 5-10 individually moving objects. We show a quantitative comparison with a video prediction baseline PredRNN~\cite{wang2022predrnn}. As shown in \tbl{table:cater}, our model achieves better performance across all three metrics. 

\begin{figure}[t]
  \centering
  \includegraphics[width=\linewidth]{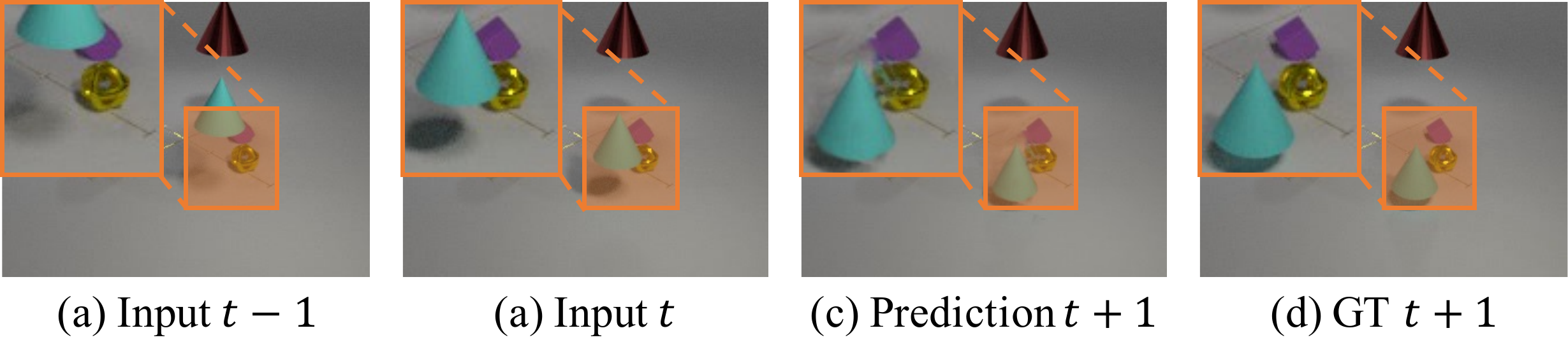}
   \caption{Model prediction on an example scene with occlusion. (a) and (b) are two historical frames as model inputs, (c) and (d) are the predicted and ground truth next frame, respectively. Top-left corners of subfigures are zoomed-in views for occluded regions. 
   }
   \label{fig:cater_appendix}
\end{figure}

Qualitatively, our method makes temporal prediction consistent with the ground truth object motions on this dataset. In \fig{fig:cater_appendix}, the model correctly recovers the purple object and the gold object occluded by the blue cone. Our model effectively handles object occlusions by warping from neighboring pixels with similar RGB values.

\subsection{Discussions}
\label{appendix:discussions}

While we focus on demonstrating the possibility of simultaneous extrapolation in both space and time, specific modules can be further optimized for each task. For example, it is possible to improve the dynamic scene representation to better handle video prediction with long horizons or highly complex motion, or to synthesize novel views with a large viewpoint change.

In the meantime, while our method is designed for natural scenes with many potential positive impacts such as interactive scene exploration for family entertainment, like all other visual content generation methods, our method might be exploited by malicious users with potential negative impacts. We expect such impacts to be minimal as our method is not designed to work with human videos. In our code release, we will explicitly specify allowable uses of our system with appropriate licenses. We will use techniques such as watermarking to label visual content generated by our system.

\end{document}